\documentclass{article}

\usepackage[preprint]{neurips_2019}

\usepackage[utf8]{inputenc} 
\usepackage[T1]{fontenc}    
\usepackage{hyperref}       
\usepackage{url}            
\usepackage{booktabs}       
\usepackage{amsfonts}       
\usepackage{nicefrac}       
\usepackage{microtype}      

\usepackage{amsmath}
\usepackage{bm}
\usepackage{booktabs}
\usepackage{graphicx}
\usepackage{soul}
\usepackage{footnote}
\title{Unsupervised Representation for EHR Signals and Codes as Patient Status Vector}
\author{%
    Sajad Darabi, Mohammad Kachuee, Majid Sarrafzadeh\\
    University of California Los Angeles\\
    \texttt{\{sajad.darabi, mkachuee, majid\}@cs.ucla.edu}
}

\begin{document}

\maketitle

\begin{abstract}
Effective modeling of electronic health records presents many challenges as they contain large amounts of irregularity most of which are due to the varying procedures and diagnosis a patient may have. Despite the recent progress in machine learning, unsupervised learning remains largely at open, especially in the healthcare domain.  In this work, we present a two-step unsupervised representation learning scheme to summarize the multi-modal clinical time series consisting of signals and medical codes into a patient status vector. First, an auto-encoder step is used to reduce sparse medical codes and clinical time series into a distributed representation. Subsequently, the concatenation of the distributed representations is further fine-tuned using a forecasting task. We evaluate the usefulness of the representation on two downstream tasks: mortality and readmission. Our proposed method shows improved generalization performance for both short duration ICU visits and long duration ICU visits.
\end{abstract}

\section{Introduction}
\noindent Learning patient representation is a popular topic in health analytics. With the availability of electronic health records (EHR) systems, it has opened avenues in learning these representations using deep learning methods. The general approach to learning representations is through the use of supervised signals. On the other hand, unsupervised learning can leverage data regardless of the presence of labels. As healthcare naturally suffers from limited labels and high costs with obtaining labels, learning reusable feature representations from large unlabeled samples has been an area of active research.  Unsupervised learning can be used to produce representations of general utility, that can be used further for downstream tasks such as mortality, readmission, length of stay, etc.

\begin{figure}[h]
    \centering
    \includegraphics{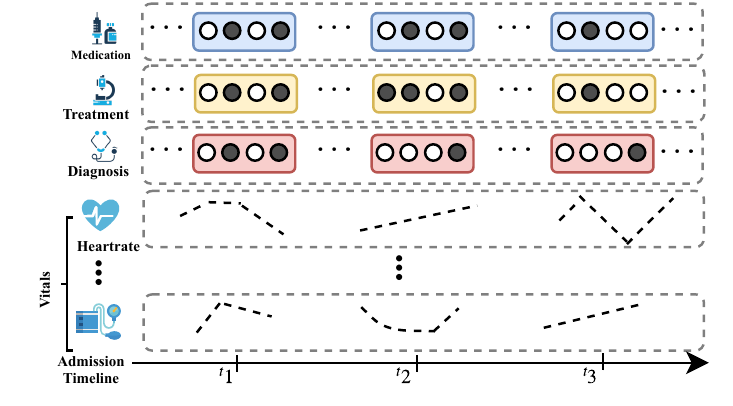}
    \caption{EHR patient timeline contains multi-modal sparse data each of which are entered at varying frequencies.}
    \label{fig:timeline}
\end{figure}

Routine medical practice generates a wealth of patient time series most of which are preempted by conditions a patient may have as they undergo care. Annotating these different data inputs to the system often require medical experts incurring large costs to label the data. Further, applying machine learning methods on patient data is not immediately obvious due to complicating factors in the collection process. First, each patient could contain varying different types such as vitals, text, and code entered into their record i.e. the data is multimodal (Fig. \ref{fig:timeline}). Second, in the recorded values there are many missing and incomplete values requiring pre-processing or imputation techniques. Lastly, there is a natural structure in the collection process that may not be captured in a single patient, but more so when looking at a body of patients. 

The use of unsupervised methods for learning representations is widely recognized for solving problems with limited data and label information. Perhaps the most common are autoencoder architectures that attempt to map the input to the same input with minimal distortion \citep{baldi2012autoencoders,bengio2013representation}. Similar ideas are commonly employed in natural language processing tasks which learn word embeddings \citep{bengio2003neural,mikolov2013}. Recently, these methods have been employed to learn medical concepts and have shown promising results \citep{choi2016multi,Choi:2017:GGA:3097983.3098126,cai2018medical,darabi2019taper}.

Previous work at large use unsupervised learning solely on clinical concepts, and disregard the patients progress throughout their ICU visit. It is not obvious how to combine other data modes to obtain a patient representation. Also, as embeddings are made to predict the surrounding context, the data used to train the embeddings are limited to patients with multiple visits which could drastically reduce the dataset size. In this paper, we propose to use unsupervised learning on both clinical concepts and vital signals using networks that take into account the sequential context of a patient. The obtained representations can be later used as a patients status vector for downstream task prediction.


The contribution of the paper are the following: 

\begin{itemize}
    \item We propose a two step unsupervised fine tuning task for embedding patient data: (1) a single visit autoencoding step followed by a (2) forecasting autoencoding step
    \item We use multi-modal data, namely, clinical concepts $+$ vital signals and show improved generalization performance for both long duration and short duration ICU stays on eICU dataset.
\end{itemize}

\section{Related Work}
The unsupervised learning used in this work stems from the widely used auto-encoding principle \citep{rumelhart1988learning,baldi2012autoencoders}. Deep auto-encoder architectures are pre-trained and followed by a supervised learning phase to train a top classifier layer that is fine-tuned on specific tasks \citep{hinton2006reducing,hinton2006fast}. These have been naturally extended to other architectures and learning methods such as variational, denoising, convolutional autoencoders\citep{vincent2008extracting,masci2011stacked,kingma2013auto}. In \cite{mikolov2013} they extended this idea to natural language processing and introduced two intuitive methods for word embeddings:  (1) skip-gram where a current word is said to predict surrounding words, (2) continuous bag of words (CBOW) where a set of words are made to predict a center word  \citep{bahdanau2014neural}. Naturally, they were extended to sequence models which are more fitting for sentences and learning sequential dependencies. For example, the popular Seq2Seq architecture is commonly used to learn language models for language generation \citep{sutskever2014sequence}. As these methods have become widespread in NLP they have also been extended to medical settings

Unsupervised learning techniques have been successfully applied to medical concepts. For instance, in \cite{miotto2016deep}, an autoencoder multi-layer perceptron was used to predict future disease codes. Similarly, in \citep{choi2016multi,suresh2017use}, a multi-layer perceptron using the skip-gram model was trained with an added regularization for the co-occurrence of codes within a visit. As EHR data might have limited occurrences of certain codes and hence hinder the learning process of deep models, later work suggested to add an attention mechanism on external ontology's \citep{Choi:2017:GGA:3097983.3098126}. More recent work on clinical concept embedding focused on leveraging temporal context using attention \citep{cai2018medical}. Others embed a structural prior into the learning process \citep{MIME2018}, where medication and procedure codes are made to predict diagnosis codes enabling the model to benefit from the structure present in EHRs. These methods are mostly limited to clinical concepts and do not use other portions of EHR data such as vital signals.

Few works have studied unsupervised methods for clinical time series \citep{kachuee2018ecg}. In \cite{lyu2018improving}, they use a Seq2Seq model to representations of clinical time series with an autoencoder loss. Following this, they assess the representation by feeding them to an LSTM classifier for downstream tasks. They show the benefits of unsupervised learning for both limited data settings and prediction performance. The majority of previous work have studied clinical time series and medical concepts independently. In this paper, we focus on an unsupervised framework for learning a patient status vector using these data streams as inputs.

\label{sec:proposed}
\section{Proposed Method}
\begin{figure}
    \centering
    \includegraphics[width=\linewidth]{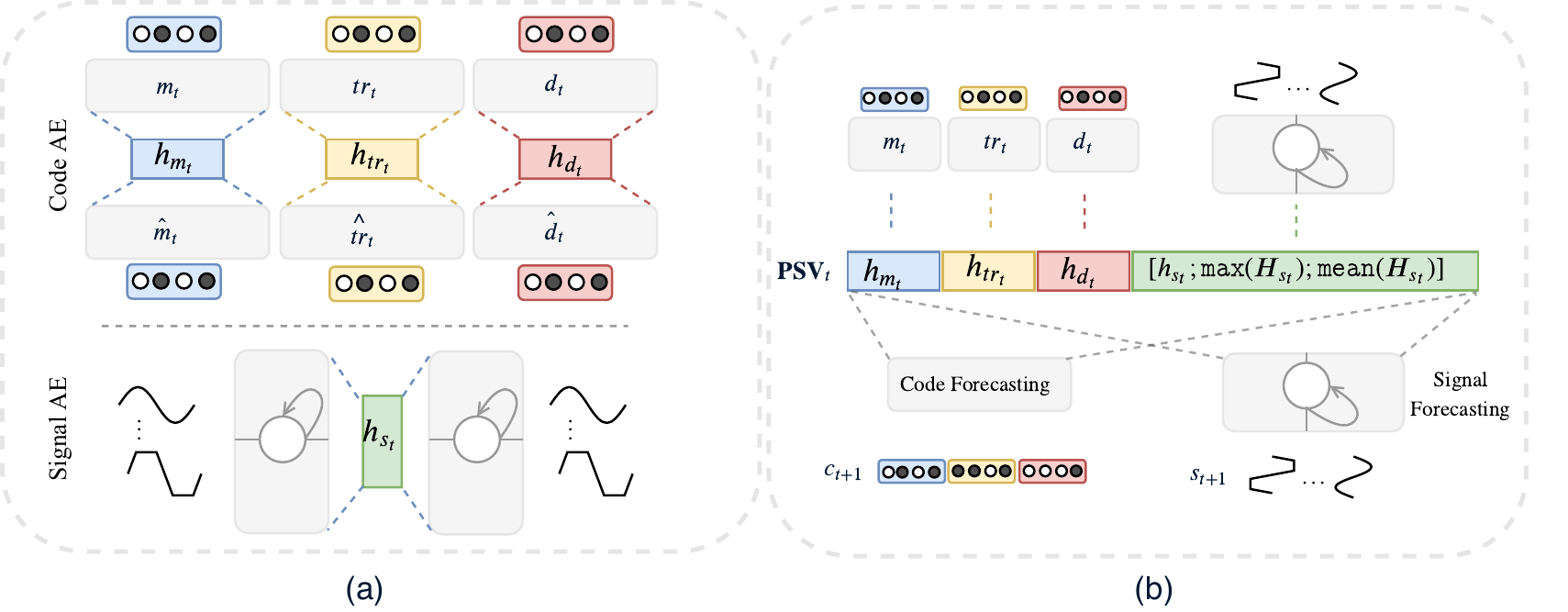}
    \caption{Overview of the proposed method: (a) autoencoding step for each datastream, data fed at this stage is predicting data occuring in the same stay. (b) forecasting step using the concatenation of the embeddings ($\textbf{PSV}_t$). The next time step is predicted in this step. }
    \label{fig:overview}
\end{figure}

In the current presentation, we are interested in learning a patient embedding using both clinical time-series signals and concepts. The usefulness of the embedding is then evaluated on two downstream tasks: in-hospital mortality, and readmission. The overview of our method is a two-step unsupervised task (Figure. \ref{fig:overview}): the first task is an auto-encoding task, and the second task is a unified forecasting task using the concatenation of the learned intermediate representations.

To introduce the problem setting, consider a sequence of EHR data that consists of finite number of medical concepts $\mathcal{C} = \mathcal{M} \cup \mathcal{D} \cup \mathcal{T}$, where $\mathcal{M}$ is the set of medication codes, $\mathcal{D}$ is the set of diagnosis codes, and $\mathcal{T}$ is the set of treatment codes. Additionally, a set of clinical time-series $\mathcal{S}$ are measured at varying frequencies. 

The EHR data of a patient may contain multiple hospital admissions each of which could contain multiple visits to the ICU. As will be explained in the data section there isn't enough information provided in the data used in this paper to distinguish the order of hospital visits for a patient. As a result, we treat each admission independent of past hospital admissions and hence timestamps are limited to the hospital admission. Following this then the longitudinal patient data for a hospital visit can be written as $D_{n} = \{\mathcal{I}_1, \mathcal{I}_2 \cdots \mathcal{I}_n\}$, where each $\mathcal{I}_i$ denotes an ICU visit with varying duration. A visit contains the data sequence $\mathcal{I}_i^T = \{(\textit{m}_1, \textit{d}_1, \textit{tr}_1, \textit{s}_1), (\textit{m}_2, \textit{d}_2, \textit{tr}_2, \textit{s}_2) \cdots (\textit{m}_T, \textit{d}_T, \textit{tr}_T, \textit{s}_T)\}$, where formally time units in an ICU visit can be defined as minutes, hours or days.

\subsection{Unsupervised Autoencoder}
Autoencoders models contain an encoding portion followed by a decoder. The encoding maps the input $x \in \Re^d$ to an intermediate representation $\Re^m$. Typically $m$ is chosen to be less than $d$, and is called a bottleneck mechanism forcing the network to learn a useful representation. The decoder is then tasked to map the representation back to the original data space $\Re^d$.

If we let $f_\theta$ and $g_\theta$ denote the encoder and decoder block respectively, the objective of the autoencoder can be written as follows:

$$
    \bm{e} = f_{\theta}(x) , \hat{x} = g_{\phi}(e)
$$
$$\mathcal{L} (x, \hat{x}) = \frac{1}{N}\sum_i^N||x - \hat{x}||_2^2$$

The MSE loss is used for real valued reconstruction whereas for multiclass one-hot vectors the cross-entropy loss is better suited. Ideally, a good performing autoencoder given sufficient context should reconstruct the input with minimal distortion.

\subsubsection{Medical Concept Embedding}
We use an autoencoder to embed single visits in ICU patients. As medical concepts are time stamped in an ICU visit it is important to take into account the sequential dependencies as well. To this end, we use transformer networks to learn the embeddings \citep{vaswani2017attention}. Transformers are known to be suited for sparse sequential data and learning inherent structures present by using self-attention. Each concept type is fed to its network with an autoencoder loss (Figure. \ref{fig:overview}a). As we consider three code types, three transformer networks are trained independently on the respective code inputs. More concretely, given a sequence of diagnosis codes $\{d_1, d_2 \cdots d_T \}$ the objective of each network is to predict the input following an autoencoder bottleneck.

$$L(d_t, \hat{d_t}) = -\sum_i d_i log(\hat{d_i}) + (1 - d_i) log(1 - \hat{d_i})$$

where $\hat{d}_i$ is the predicted set of diagnosis codes following the bottleneck.

\subsubsection{Clinical Time Series Embedding} While Seq2Seq models are commonly used for supervised tasks and language modeling, we use it to embed clinical time series of a patient into a representation by minimizing the reconstruction error. GRU cells, a recurrent neural network variant, are used as the basic block of the encoder and decoder of the Seq2Seq model.  The input to this network is the sequence of time series accompanying clinical concepts denoted as $\{s_1, s_2 \cdots s_T \}$, where each sample $s_t$ is a feature vector $\Re^{|S|}$. As noted earlier, patients may have missing attributes at each time step, to this effect we present a mask vector $m_t \in \{0, 1\}^{|S|}$, where a 1 presents the feature is observed. The Seq2Seq model is then trained by windowing the clinical time series with window size $w$ and setting the objective to the MSE autoencoder loss. The objective is slightly modified to penalize predictions when there is an observed value at the corresponding time step. Formally written as 
\begin{align*}
    \mathcal{L}(S_{1:T}, \hat{S}_{1:T}) = \frac{1}{N}\sum_i^{\frac{T}{w}} &m_{wk:w(k+1)}\\ \cdot &||s_{wk:w(k+1)} - \hat{s}_{wk:w(k+1)}||_2^2
\end{align*}

\subsection{Unsupervised Forecasting Task}
Following the autoencoder unsupervised task on each component, we combine the representations and further fine-tune on a forecasting task, that is, predicting the next visits diagnosis, medication, treatment codes, and vitals (Figure.\ref{fig:overview} b).

To combine the learned diagnosis, medication, treatment and clinical time series we simply concatenate the representations to create a $\text{PSV}_t = [e_{d_t};e_{m_t};e_{tr_t};e_{s_t}]$ at time $t$, which we call the patient status vector (PSV). 

Each code representation is obtained by embedding the set of codes until time $t$ and taking the final hidden representation as the representation of the code. For example, the set of diagnosis codes $\{d_1, \cdots d_t\}$ are embedded using the transformer network creating a set of representations $H_d = \{h_{d_1} \cdots h_{d_t}\}$. Then the final hidden representation $e_{d_t} = h_{d_t}$ is used.  

On the other hand as there are many more data points for clinical time series, the accumulated vitals until time $t$ are first windowed using the same window size $w$ in the autoencoding step. The resulting is a set of sequences $\{s_{1:w}, s_{w:2w}, \cdots s_{w(\frac{t}{w} - 1):t} \}$. Subsequently, a hidden representation is obtained from the encoder cell in the Seq2Seq model $H_s = \{h_{s_1}, \cdots , h_{s_{t}}\}$ and the final representation is obtained as

$$e_{s_t} = [h_{s_t}; \texttt{maxpool}(H_s); \texttt{meanpool}(H_s)]$$

where the components are concatenated. This is done as the time-series signals could get very length and information may get lost if we were to only consider the final representation of the time series. Intuitively by taking the max-pool, we are looking at a part of the signal resulting in high activation, the mean-pool a baseline measure and the final representation as to the most recent condition of the patients signal.

Additionally, we add the patients demographics $z_t \in \Re^{|Z|}$ containing, age, weight, height, and gender where discrete values are coded as one hot vectors. The final representation is obtain as
$$\text{PSV}_t = [e_{d_t};e_{m_t};e_{tr_t};e_{s_t}; z_t]$$

This representation is used for the forecasting task: 1) predicting the set of codes at the next time step and 2) predicting the set of vitals for the next time window. The objective of the forecasting unsupervised step can be written as

\begin{align*}
&\frac{1}{2T}\sum_{t=1}^T\sum_{0\leq j\leq w, j\neq0} log(p(c_{t+j}|\text{psv}_t)) +\\
&\frac{1}{2\frac{T}{w}}\sum_i^{\frac{T}{w}} m_{wk:w(k+1)}\cdot ||s_{wk:w(k+1)} - \hat{s}_{wk:w(k+1)}||_2^2
\end{align*}

where a fully connected layer followed by softmax is used to model $p(c_{t+j}|\text{PSV}_t)$. On the other hand, $\hat{s_t}$ is a GRU decoder model with hidden input as the concatenation of medical code representations along with the generated encoder representations. 

Based on the suggested representation learning scheme, the trained networks are used to embed patients given a medical record and later fed to a classifier for downstream tasks. Training details and hyperparameters used for each specific task are detailed in the experiment section.


\section{Experiments}
\subsection{Data}
The dataset used is the publicly available eICU collaborative Research Database v2.0 \citep{pollard2018eicu}. The eICU consists of over 200,000 Intensive Care Unit (ICU) records collected from over 250 hospital sites in the United States, between 2014 and 2015. The data is indexed through unique patient identifiers, where each patient could have multiple hospital admissions and within each hospital admission they could be admitted to an ICU multiple times. Timestamps in a patients record are referenced from the ICU admission stay, in which patients could have certain delay before being admitted to an ICU. As the tables provided in the data do not allow one to order hospital admissions the downstream tasks are limited to within hospital admissions, and further the patients history is limited to the visit as well. From the encounter records, we extract diagnosis codes, medication codes, and treatment codes. Additionally, we extract periodic vital signals collected regularly by bedside monitors. 

\subsubsection{Preprocessing} Encounters in the ICU vary greatly from patient to patient as a result we preprocess the dataset to include sufficient examples for learning. Patients younger than 16 years of age are removed. The dataset contains patients with burns and organ donors or admission after a transplant which are also removed from the dataset. The periodic bedside monitoring devices are down-sampled to median values every 5 minutes by the original creators of the dataset; We further down-sample by binning the values every 60 minutes (12 samples). The median is used for each bin. Following the pre-processing steps, we normalize signals and patient demographics to have zero mean and unit variance. We summarize the statistics of the compiled cohort in Table \ref{tab:Cohort}. Two cohorts are considered of different lengths: short ICU visits of 1 hour to 24 hours, and long ICU visits of 24 hours to 720 hours duration. Our preprocessing steps closely resemble that of APACHE IV \citep{zimmerman2006acute} a gold-standard metric in which we have used similar exclusion criteria. 

\begin{table}[h]
\caption{Cohort Statistics of Compiled Data}
\centering
\label{tab:Cohort}
\renewcommand{\arraystretch}{1.3}
\begin{tabular}{l cccc}
    \toprule
    \multicolumn{1}{c}{}  & \multicolumn{1}{c}{1h-24h}&\multicolumn{1}{c}{24h-720h}
    \\ \hline
     \# hospital visits & 44190 & 95649\\
     \# ICU stays &  46664 & 110270\\
    \# of diagnosis codes (avg/visit) & 918 (1.17) & 918 (3.075) \\
    \# of medication codes (avg/visit) & 1412 (8.52) & 1412 (21.66)   \\
    \# of treatment codes (avg/visit) & 2711 (1.12) & 2711 (2.85)  \\
      length of signals (avg, min, max)&  (14.6, 0, 205) & (87, 0, 832) \\
    \# of in-hospital mortality & 3249 & 6157\\
    \# of within visit ICU readmissions & 10858 & 21742\\
    \bottomrule
\end{tabular}
\end{table}

\subsection{Downstream tasks}
To evaluate the usefulness of the learned representation we train a classifier on top of the representation on downstream tasks and evaluate its generalization performance. 

\textbf{Mortality}: Given a patients record and patients history predict the patient's death during the ICU stay. This is a binary prediction task.

\textbf{Readmission} Similarly, in this task we focus on predicting whether the patient will be readmitted to ICU again within the same visit.

\subsection{Training Details \& Evaluation  \& Baselines}
The unsupervised training follows the method described in Section \ref{sec:proposed}. 

The medical concept autoencoder blocks are transformer layers with multi-head attention as described in \citep{vaswani2017attention}. This block contains multiple hyperparameters namely number of multi-head attention heads $n_{head}$, dimension for each head $d_{head}$ and the final model representations $d_{code}$. We set these to 8, 64 and 256 respectively for all three medical concept embedding blocks 2 such layers are stacked. The network is trained using Adam \cite{kingma2014adam} with a cosine annealing schedule and a period of 50 epochs. The initial learning rate is set to $0.00025$. 

The Seq2Seq model is trained by setting the window size for clinical signals to 24 (i.e. a full day). Missing values are imputed with the median of the statistics of the complete dataset if the patient does not have any history for the vital; otherwise, the patients average is used to impute the corresponding missing signal. The encoder and decoder blocks are bidirectional GRU cells where the encoder hidden layer dimension is set to $d_{enc} = 128$ and the decoder is set to $256$. A step learning rate with initial lr set to $0.001$ and step decay rate of 50 is used to train the network for 100 epochs.

Following the autoencoding training, the dataset is filtered to contain visits with multiple ICU stays for the forecasting step. To avoid catastrophic forgetting we gradually unfreeze the blocks and allow them to train for 2 epochs \citep{howard2018universal}. This is similar to chain thaw proposed in \cite{felbo2017using}, where each layer is trained at a time. Though here we gradually unfreeze the whole model and allow it to learn altogether. Our final patient status vector representation is of dimension 1762 fed to a two-layer fully connected layer with ReLU as activation and dropout set to 0.1 in between layers. 

We implemented all the models with Pytorch 1.0 \cite{paszke2017automatic}. For training the models we use the Adam optimizer \cite{kingma2014adam}.  In all experiments, the batch size is set to 64.

\subsubsection{Evaluation}
For all experiments, we use $15\%$ of the data selected randomly as the test set. The remainder is used for training/validation splits. In all experiments, the presented values are the average of 5 experiments for each task.\\

\textbf{Area under the precision-recall (AU-PR)}: this metric is the cumulative area under the curve by plotting precision and recall while varying the outputs $P(y=1|(c_t, t_t))$ true/false threshold from 0 to 1.\\
\textbf{Receiver operating characteristic curve (AU-ROC)}: This metric is the area under the plot of the true positive rate against false positive rate while varying  outputs $P(y=1|(c_t, t_t))$ true/false threshold from 0 to 1.
\subsubsection{Baseline Models}
\begin{itemize}
    \item \textbf{Transformer Embedding}: The set of medical concepts in the records are embedded by training a transformer network using the skip-gram model. This representation is used as a patient representation for downstream tasks.
    \item \textbf{Seq2Seq} (Unsupervised): A Seq2Seq model is trained on clinical times series portion of the data, which can then be used to embed patients signals for the downstream task. 
    \item \textbf{Seq2Seq} (Semi-supervised): A Seq2Seq model is trained on clinical times series portion of the data, and further fine-tuned on downstream tasks.
\end{itemize}

We also perform an ablation on different components of the patient status vector by including/excluding portions of the representations



\section{Results}

To evaluate the PSV representation we consider both a short duration cohort and long duration cohort. This is done as the dynamics of short term ICU stays and long term ICU stays vary greatly. The results for both downstream tasks are summarized in Table \ref{tab:mort_comparisoin}\&\ref{tab:red_comparison}. 

In the tables, the ablation of the different components of the PSV model is done by first pre-training the model in an unsupervised fashion and freezing the pre-trained network for downstream tasks. Empirically, from both tables, it is evident the network benefits from both code and signal representations on the defined tasks. 

The baseline models use only portions of the complete data under consideration and to make a fair comparison we can compare the respective components of the PSV model reported accordingly. For example, Seq2Seq \citep{lyu2018improving} can be compared to {PSV} {\small \texttt{(Signal)}}. Similarly, Transformer {\small \citep{darabi2019taper}} can be compared to {PSV} {\small \texttt{(Code)}}. In both cases, the difference is largely in the unsupervised training step. By leveraging both single visit ICU data and multi-visit ICU patients we can achieve a better initialization for downstream tasks.

Lastly, from the results, the prediction performance drops (significantly in some cases) when comparing short visits and long visits. This could largely be due to dynamics that are present in longer ICU visits or human factors that are not present/captured in the dataset.
\begin{table*}[!h]
\caption{Mortality downstream task ablation, and comparison with baselines. Results reported are average of 5 runs on a test split of $15\%$ and the std are reported in parenthesis.}
\label{tab:mort_comparisoin}
\begin{center}
\renewcommand{\arraystretch}{1.5}
\resizebox{\columnwidth}{!}{%
\begin{tabular}{l|cc|cc}
\toprule
 \multicolumn{1}{c}{\bf Dataset {\small (\texttt{Mortality task})}} &\multicolumn{2}{c}{\bf 1h-24h} &\multicolumn{2}{c}{\bf 24h-720h} 
\\ \hline 
& PR-AUC & ROC & PR-AUC & ROC\\\hline
 {PSV} {\small \texttt{(Code+Signal)}} & 62.46  {\footnotesize ($\pm$ 0.20)} & \bf 90.06  {\footnotesize ($\pm$ 0.12)} & 48.88  {\footnotesize ($\pm$ 0.13)} & \bf 85.90  {\footnotesize ($\pm$ 0.09)}  \\
{PSV} {\small \texttt{(Code)}}   &  45.35 {\footnotesize ($\pm$ 0.22)} &  81.69 {\footnotesize ($\pm$ 0.29)} & 30.50 {\footnotesize ($\pm$ 0.22)} & 77.98 {\footnotesize ($\pm$ 0.14)}  \\
{PSV} {\small \texttt{(Signal)}} &  49.42 {\footnotesize ($\pm$ 0.18)}  &  82.27 {\footnotesize ($\pm$ 0.04)} &  31.32 ({\footnotesize $\pm0$.10)}  & 82.27 {\footnotesize ($\pm$ 0.04)}  \\
{PSV} {\small \texttt{(Semi-supervised)}} & \bf 65.40  {\footnotesize ($\pm$ 0.35)} &  89.10  {\footnotesize ($\pm$ 0.59)} &  \bf 53.76  {\footnotesize ($\pm$ 0.26)} & 85.15  {\footnotesize ($\pm$ 0.66)}  \\\hline
Seq2Seq {\small \citep{lyu2018improving}}\footnotemark& 7.73  {\footnotesize ($\pm$ 0.60)}   & 51.32 {\footnotesize ($\pm$ 0.31)} & 8.17  {\footnotesize ($\pm$ 0.05)} & 61.90  {\footnotesize ($\pm$ 0.19)} \\
Seq2Seq {\small \texttt{(Semi-supervised)}} &   8.58{\footnotesize ($\pm$0.32)}  &   51.23{\footnotesize ($\pm$ 0.22)}  & 19.22 {\footnotesize ($\pm$ 0.06)} &  61.63 {\footnotesize ($\pm$ 61.63)}  \\
Transformer {\small \citep{darabi2019taper}}$^2$  & 11.18 {\footnotesize ($\pm$ 0.35)} & 53.01  {\footnotesize ($\pm$ 0.64)} &  10.67 {\footnotesize ($\pm$ 0.30)} &  50.45  {\footnotesize ($\pm$ 0.88)} \\
\bottomrule
\end{tabular}%
}
\end{center}
\end{table*}
\footnotetext{Results reported using our re-implementation, further the input are limited to the same set of vitals/codes under consideration.}

\begin{table*}[!h]
\caption{Readmission downstream task ablation, and comparison with baselines. Results reported are average of 5 runs on a test split of $15\%$ and the std are reported in parenthesis.}
\label{tab:red_comparison}
\begin{center}
\renewcommand{\arraystretch}{1.5}
\resizebox{\columnwidth}{!}{%
\begin{tabular}{l|cc|cc}
\toprule
 \multicolumn{1}{c}{\bf Dataset {\small (\texttt{Readmission task})}} &\multicolumn{2}{c}{\bf 1h-24h} &\multicolumn{2}{c}{\bf 24h-720h} 
\\ \hline 
& PR-AUC & ROC & PR-AUC & ROC\\\hline
 {PSV} {\small \texttt{(Code+Signal)}} & 61.25  {\footnotesize ($\pm$ 0.26)} & 80.99  {\footnotesize ($\pm$ 0.11)} & 57.86  {\footnotesize ($\pm$ 0.19)} & 80.94  {\footnotesize ($\pm$ 0.09)}  \\
{PSV} {\small \texttt{(Code)}}   &  57.24 {\footnotesize ($\pm$ 0.56)} &  79.58 {\footnotesize ($\pm$ 0.12)} &  49.63 {\footnotesize ($\pm$ 0.38)}  &  76.15  {\footnotesize ($\pm$ 0.24)}  \\
{PSV} {\small \texttt{(Signal)}} &  30.47 {\footnotesize ($\pm$ 0.13)} &  59.35 {\footnotesize ($\pm$ 0.15)} & 30.34  {\footnotesize ($\pm$ 0.10)} &  59.22 {\footnotesize ($\pm$ 0.14)}  \\
{PSV} {\small \texttt{(Semi-supervised)}} & \bf 69.02  {\footnotesize ($\pm$ 0.42)} &  \bf 83.40 {\footnotesize ($\pm$ 0.23)} &  \bf 68.04 {\footnotesize ($\pm$ 0.51)} &  \bf 82.25 {\footnotesize ($\pm$ 0.24)}  \\\hline
Seq2Seq {\small \citep{lyu2018improving}}$^2$ & 26.43  {\footnotesize ($\pm$ 0.56)}  &  51.45  {\footnotesize ($\pm$ 1.13)} & 20.30  {\footnotesize ($\pm$ 0.18)} & 52.35  {\footnotesize ($\pm$ 0.34)} \\
    Seq2Seq {\small \texttt{(Semi-supervised)}} &   26.68{\footnotesize ($\pm$ 0.19)}  &  51.80 {\footnotesize ($\pm$0.54)}  & 22.31 {\footnotesize ($\pm$ 0.11)} &  52.18 {\footnotesize ($\pm$ 0.42)}  \\
Transformer {\small \citep{darabi2019taper}}$^2$ & 28.22 {\footnotesize ($\pm$ 0.60)} &  58.82  {\footnotesize ($\pm$ 0.43)} &  27.70 {\footnotesize ($\pm$ 0.79)} &  59.45 {\footnotesize ($\pm$ 0.91)}  \\
\bottomrule
\end{tabular}%
}
\end{center}
\end{table*}


\section{Conclusion}
In this paper, we have introduced an unsupervised patient status vector embedding scheme for EHR patient longitudinal data. The method effectively leverages both single-visit ICU patients and multi-visit ICU patients using a two-step autoencoding step. We have evaluated the proposed method using two cohorts compiled from the eICU EHR dataset of different duration by using periodic vital signals and medical codes as input to our model. From empirical results on downstream tasks, the proposed unsupervised learning approach outperforms previous work. Lastly, presented in the results, long-stay ICU visit patients present a bigger challenge for modeling; as EHR do not necessarily contain human factors that could play in the prognosis of a patient, future work could leverage different modes of data which were not considered in this work.

\bibliographystyle{plainnat}
\bibliography{bib.bib}

\begin{thebibliography}{28}
\providecommand{\natexlab}[1]{#1}
\providecommand{\url}[1]{\texttt{#1}}
\expandafter\ifx\csname urlstyle\endcsname\relax
  \providecommand{\doi}[1]{doi: #1}\else
  \providecommand{\doi}{doi: \begingroup \urlstyle{rm}\Url}\fi

\bibitem[Bahdanau et~al.(2014)Bahdanau, Cho, and Bengio]{bahdanau2014neural}
Dzmitry Bahdanau, Kyunghyun Cho, and Yoshua Bengio.
\newblock Neural machine translation by jointly learning to align and
  translate.
\newblock \emph{arXiv preprint arXiv:1409.0473}, 2014.

\bibitem[Baldi(2012)]{baldi2012autoencoders}
Pierre Baldi.
\newblock Autoencoders, unsupervised learning, and deep architectures.
\newblock In \emph{Proceedings of ICML workshop on unsupervised and transfer
  learning}, pages 37--49, 2012.

\bibitem[Bengio et~al.(2003)Bengio, Ducharme, Vincent, and
  Jauvin]{bengio2003neural}
Yoshua Bengio, R{\'e}jean Ducharme, Pascal Vincent, and Christian Jauvin.
\newblock A neural probabilistic language model.
\newblock \emph{Journal of machine learning research}, 3\penalty0
  (Feb):\penalty0 1137--1155, 2003.

\bibitem[Bengio et~al.(2013)Bengio, Courville, and
  Vincent]{bengio2013representation}
Yoshua Bengio, Aaron Courville, and Pascal Vincent.
\newblock Representation learning: A review and new perspectives.
\newblock \emph{IEEE transactions on pattern analysis and machine
  intelligence}, 35\penalty0 (8):\penalty0 1798--1828, 2013.

\bibitem[Cai et~al.(2018)Cai, Gao, Ngiam, Ooi, Zhang, and Yuan]{cai2018medical}
Xiangrui Cai, Jinyang Gao, Kee~Yuan Ngiam, Beng~Chin Ooi, Ying Zhang, and
  Xiaojie Yuan.
\newblock Medical concept embedding with time-aware attention.
\newblock \emph{arXiv preprint arXiv:1806.02873}, 2018.

\bibitem[Choi et~al.(2016)Choi, Bahadori, Searles, Coffey, Thompson, Bost,
  Tejedor-Sojo, and Sun]{choi2016multi}
Edward Choi, Mohammad~Taha Bahadori, Elizabeth Searles, Catherine Coffey,
  Michael Thompson, James Bost, Javier Tejedor-Sojo, and Jimeng Sun.
\newblock Multi-layer representation learning for medical concepts.
\newblock In \emph{Proceedings of the 22nd ACM SIGKDD International Conference
  on Knowledge Discovery and Data Mining}, pages 1495--1504. ACM, 2016.

\bibitem[Choi et~al.(2017)Choi, Bahadori, Song, Stewart, and
  Sun]{Choi:2017:GGA:3097983.3098126}
Edward Choi, Mohammad~Taha Bahadori, Le~Song, Walter~F. Stewart, and Jimeng
  Sun.
\newblock Gram: Graph-based attention model for healthcare representation
  learning.
\newblock In \emph{Proceedings of the 23rd ACM SIGKDD International Conference
  on Knowledge Discovery and Data Mining}, KDD '17, pages 787--795, New York,
  NY, USA, 2017. ACM.
\newblock ISBN 978-1-4503-4887-4.
\newblock \doi{10.1145/3097983.3098126}.
\newblock URL \url{http://doi.acm.org/10.1145/3097983.3098126}.

\bibitem[Choi et~al.(2018)Choi, Xiao, Stewart, and Sun]{MIME2018}
Edward Choi, Cao Xiao, Walter Stewart, and Jimeng Sun.
\newblock Mime: Multilevel medical embedding of electronic health records for
  predictive healthcare.
\newblock In S.~Bengio, H.~Wallach, H.~Larochelle, K.~Grauman, N.~Cesa-Bianchi,
  and R.~Garnett, editors, \emph{Advances in Neural Information Processing
  Systems 31}, pages 4547--4557. Curran Associates, Inc., 2018.

\bibitem[Darabi et~al.(2019)Darabi, Kachuee, Fazeli, and
  Sarrafzadeh]{darabi2019taper}
Sajad Darabi, Mohammad Kachuee, Shayan Fazeli, and Majid Sarrafzadeh.
\newblock Taper: Time-aware patient ehr representation.
\newblock \emph{arXiv preprint arXiv:1908.03971}, 2019.

\bibitem[Felbo et~al.(2017)Felbo, Mislove, S{\o}gaard, Rahwan, and
  Lehmann]{felbo2017using}
Bjarke Felbo, Alan Mislove, Anders S{\o}gaard, Iyad Rahwan, and Sune Lehmann.
\newblock Using millions of emoji occurrences to learn any-domain
  representations for detecting sentiment, emotion and sarcasm.
\newblock \emph{arXiv preprint arXiv:1708.00524}, 2017.

\bibitem[Hinton and Salakhutdinov(2006)]{hinton2006reducing}
Geoffrey~E Hinton and Ruslan~R Salakhutdinov.
\newblock Reducing the dimensionality of data with neural networks.
\newblock \emph{science}, 313\penalty0 (5786):\penalty0 504--507, 2006.

\bibitem[Hinton et~al.(2006)Hinton, Osindero, and Teh]{hinton2006fast}
Geoffrey~E Hinton, Simon Osindero, and Yee-Whye Teh.
\newblock A fast learning algorithm for deep belief nets.
\newblock \emph{Neural computation}, 18\penalty0 (7):\penalty0 1527--1554,
  2006.

\bibitem[Howard and Ruder(2018)]{howard2018universal}
Jeremy Howard and Sebastian Ruder.
\newblock Universal language model fine-tuning for text classification.
\newblock \emph{arXiv preprint arXiv:1801.06146}, 2018.

\bibitem[Kachuee et~al.(2018)Kachuee, Fazeli, and Sarrafzadeh]{kachuee2018ecg}
Mohammad Kachuee, Shayan Fazeli, and Majid Sarrafzadeh.
\newblock Ecg heartbeat classification: A deep transferable representation.
\newblock In \emph{2018 IEEE International Conference on Healthcare Informatics
  (ICHI)}, pages 443--444. IEEE, 2018.

\bibitem[Kingma and Ba(2014)]{kingma2014adam}
Diederik~P Kingma and Jimmy Ba.
\newblock Adam: A method for stochastic optimization.
\newblock \emph{arXiv preprint arXiv:1412.6980}, 2014.

\bibitem[Kingma and Welling(2013)]{kingma2013auto}
Diederik~P Kingma and Max Welling.
\newblock Auto-encoding variational bayes.
\newblock \emph{arXiv preprint arXiv:1312.6114}, 2013.

\bibitem[Lyu et~al.(2018)Lyu, H{\"u}ser, Hyland, Zerveas, and
  R{\"a}tsch]{lyu2018improving}
Xinrui Lyu, Matthias H{\"u}ser, Stephanie~L Hyland, George Zerveas, and Gunnar
  R{\"a}tsch.
\newblock Improving clinical predictions through unsupervised time series
  representation learning.
\newblock \emph{arXiv preprint arXiv:1812.00490}, 2018.

\bibitem[Masci et~al.(2011)Masci, Meier, Cire{\c{s}}an, and
  Schmidhuber]{masci2011stacked}
Jonathan Masci, Ueli Meier, Dan Cire{\c{s}}an, and J{\"u}rgen Schmidhuber.
\newblock Stacked convolutional auto-encoders for hierarchical feature
  extraction.
\newblock In \emph{International Conference on Artificial Neural Networks},
  pages 52--59. Springer, 2011.

\bibitem[Mikolov et~al.(2013)Mikolov, Sutskever, Chen, Corrado, and
  Dean]{mikolov2013}
Tomas Mikolov, Ilya Sutskever, Kai Chen, Greg~S Corrado, and Jeff Dean.
\newblock Distributed representations of words and phrases and their
  compositionality.
\newblock In C.~J.~C. Burges, L.~Bottou, M.~Welling, Z.~Ghahramani, and K.~Q.
  Weinberger, editors, \emph{Advances in Neural Information Processing Systems
  26}, pages 3111--3119. Curran Associates, Inc., 2013.

\bibitem[Miotto et~al.(2016)Miotto, Li, Kidd, and Dudley]{miotto2016deep}
Riccardo Miotto, Li~Li, Brian~A Kidd, and Joel~T Dudley.
\newblock Deep patient: an unsupervised representation to predict the future of
  patients from the electronic health records.
\newblock \emph{Scientific reports}, 6:\penalty0 26094, 2016.

\bibitem[Paszke et~al.(2017)Paszke, Gross, Chintala, Chanan, Yang, DeVito, Lin,
  Desmaison, Antiga, and Lerer]{paszke2017automatic}
Adam Paszke, Sam Gross, Soumith Chintala, Gregory Chanan, Edward Yang, Zachary
  DeVito, Zeming Lin, Alban Desmaison, Luca Antiga, and Adam Lerer.
\newblock Automatic differentiation in {PyTorch}.
\newblock In \emph{NIPS Autodiff Workshop}, 2017.

\bibitem[Pollard et~al.(2018)Pollard, Johnson, Raffa, Celi, Mark, and
  Badawi]{pollard2018eicu}
Tom~J Pollard, Alistair~EW Johnson, Jesse~D Raffa, Leo~A Celi, Roger~G Mark,
  and Omar Badawi.
\newblock The eicu collaborative research database, a freely available
  multi-center database for critical care research.
\newblock \emph{Scientific data}, 5, 2018.

\bibitem[Rumelhart et~al.(1988)Rumelhart, Hinton, Williams,
  et~al.]{rumelhart1988learning}
David~E Rumelhart, Geoffrey~E Hinton, Ronald~J Williams, et~al.
\newblock Learning representations by back-propagating errors.
\newblock \emph{Cognitive modeling}, 5\penalty0 (3):\penalty0 1, 1988.

\bibitem[Suresh et~al.(2017)Suresh, Szolovits, and Ghassemi]{suresh2017use}
Harini Suresh, Peter Szolovits, and Marzyeh Ghassemi.
\newblock The use of autoencoders for discovering patient phenotypes.
\newblock \emph{arXiv preprint arXiv:1703.07004}, 2017.

\bibitem[Sutskever et~al.(2014)Sutskever, Vinyals, and
  Le]{sutskever2014sequence}
Ilya Sutskever, Oriol Vinyals, and Quoc~V Le.
\newblock Sequence to sequence learning with neural networks.
\newblock In \emph{Advances in neural information processing systems}, pages
  3104--3112, 2014.

\bibitem[Vaswani et~al.(2017)Vaswani, Shazeer, Parmar, Uszkoreit, Jones, Gomez,
  Kaiser, and Polosukhin]{vaswani2017attention}
Ashish Vaswani, Noam Shazeer, Niki Parmar, Jakob Uszkoreit, Llion Jones,
  Aidan~N Gomez, {\L}ukasz Kaiser, and Illia Polosukhin.
\newblock Attention is all you need.
\newblock In \emph{Advances in neural information processing systems}, pages
  5998--6008, 2017.

\bibitem[Vincent et~al.(2008)Vincent, Larochelle, Bengio, and
  Manzagol]{vincent2008extracting}
Pascal Vincent, Hugo Larochelle, Yoshua Bengio, and Pierre-Antoine Manzagol.
\newblock Extracting and composing robust features with denoising autoencoders.
\newblock In \emph{Proceedings of the 25th international conference on Machine
  learning}, pages 1096--1103. ACM, 2008.

\bibitem[Zimmerman et~al.(2006)Zimmerman, Kramer, McNair, and
  Malila]{zimmerman2006acute}
Jack~E Zimmerman, Andrew~A Kramer, Douglas~S McNair, and Fern~M Malila.
\newblock Acute physiology and chronic health evaluation (apache) iv: hospital
  mortality assessment for today’s critically ill patients.
\newblock \emph{Critical care medicine}, 34\penalty0 (5):\penalty0 1297--1310,
  2006.

\end{thebibliography}

\end{document}